# Computer Poker Research at LIACC

**Luís Filipe Teófilo, Luís Paulo Reis, Henrique Lopes Cardoso, Dinis Félix, Rui Sêca, João Ferreira, Pedro Mendes, Nuno Cruz, Vitor Pereira, Nuno Passos**

LIACC – Artificial Intelligence and Computer Science Lab., University of Porto, Portugal
Rua Campo Alegre 1021 4169-007 Porto, Portugal
FEUP – Faculty of Engineering, University of Porto – DEI, Portugal
Rua Dr. Roberto Frias, s/n 4200-465 Porto, Portugal
luis.teofilo@fe.up.pt, lpreis@dsi.uminho.pt, hlc@fe.up.pt
http://www.liacc.up.pt/

**Abstract**

Computer Poker's unique characteristics present a well-suited challenge for research in artificial intelligence. For that reason, and due to the Poker's marked increase in popularity in Portugal since 2008, several member of LIACC have researched in this field.

Several works were published as papers and master theses and more recently a member of LIACC engaged on a research in this area as a Ph.D. thesis in order to develop a more extensive and in-depth work.

This paper describes the existing research in LIACC about Computer Poker, with special emphasis on the completed master's theses and plans for future work. This paper means to present a summary of the lab's work to the research community in order to encourage the exchange of ideas with other labs / individuals. LIACC hopes this will improve research in this area so as to reach the goal of creating an agent that surpasses the best human players.

## I – Introduction

LIACC members research in several areas in the field of artificial intelligence, robotics, simulation and multi agent systems. Some examples of successful projects include the Robotic Soccer team FC Portugal (several times world champion in different categories) and the Intellwheels project (an intelligent wheelchair designed to provide enhanced mobility for people with physical disabilities).

Since 2008 there has been research at LIACC about Computer Poker. This coincided with the increase in popularity of the game, especially the Texas Hold'em variant. Moreover, the unique characteristics of the game (such as the need for opponent modeling or the presence incomplete information) present a challenge that is perfectly aligned with the lab's research goals.

The lab's work in the Computer Poker domain can be found in several papers published both in national and international conferences, most of which resulted from completed master theses. Moreover, a member of LIACC recently started research in this area as a Ph.D. thesis in order to develop a more extensive and in-depth work.

The aim of this paper is the dissemination of the work done on Computer Poker by LIACC members so as to promote it and to stimulate the exchange of ideas with other researchers in the field.

The rest of the paper is organized as follows. Section II briefly describes some related work on the Computer Poker domain. Section III presents completed Poker research work done at LIACC with special emphasis on the published master theses. Section IV describes ongoing research by presenting recent developments as well as future work ideas. Finally, some conclusions are drawn in section V.

## II - Related Work

The research on Computer Poker has been active over the past 10 years, which is demonstrated by the relatively high number of publications in top conferences and journals, as well as completed master and doctoral theses.

The most relevant work in the area was done by a research group exclusively dedicated to Computer Poker, the Computer Poker Research Group (CPRG) at University Alberta.

The first approaches to build Poker agents were rule-based, which involves specifying the action that should be taken for a given game state [DP'98-1, AD'02-1]. These approaches led to the creation of the first agents that were able to defeat weak human opponents. Another important work [FD'01-1] with comparable success applied a reinforcement learning algorithm based on Q-Learning in a simplified version of Texas Hold'em. In this approach, the agent was able to learn how to play against several types of opponents.

The greatest breakthrough in Poker research so far began with the use of Nash's equilibrium theory in agents. Since then, several approaches based on Nash Equilibrium emerged: Best Response, Restricted Nash Response and data-biased response. Currently, one of the best known Poker agents – Polaris [MJ'07-1] – uses a mixture of these approaches.

Other recent methodologies were based on pattern matching [LT'11-1, AK'10-1] and on the Monte Carlo Search Tree algorithm [AK'10-1, GB'09-1].

One notable work it Darse Billings Ph.D. thesis [DB'06-1] which evaluates and compares several methodologies for agent building.

Despite all the breakthroughs achieved and to the best of the authors' knowledge there is no known approach in which the agent has consistently reached a level similar to a competent human player.

## III - Completed Research

This section briefly describes completed research works about Computer Poker that was carried out at LIACC.

### 1. Opponent Modeling in Texas Hold'em (2008)

The first research work done at LIACC on the field of Computer Poker was developed by Dinis Félix [DF'08-1] as a master thesis. The work culminated in the publication of two papers [DF'08-2, DF'08-3].

This work is focused on exploring opponent modeling methodologies in the Pre-Flop round of Texas Hold'em Poker. Only two features are used to classify the opponents: VP$IP – percentage of times that a player pays to see the Flop; Aggression Factor – the ratio between the number of raises and calls. By combining these features with the Sklansky Groups, eight different agents were implemented: Gambler, Maniac, Fish, Calling Station, Rock, Weak Tight, Fox and Ace.

After that, an Observer Agent (an agent that considers the VP$IP and the Aggression Factor of its opponents to adapt the strategy) was implemented. The strategy was based on the Effective Hand Strength Formula [DP'98-1] with a slight modification: instead of considering every possible two-card combinations of the remaining cards, it considers the possible opponent hands. For instance, a very tight player unlikely presents a hand with a very low score.

**Results**

The Observer Agent was put up against the eight developed agents. The observer outperformed every agent, especially the most passive ones. Another interesting result was the fact that the aggressive agents survive longer when playing against an observer agent.

### 2. An Intelligent Poker-Agent for Texas Hold'em (2008)

This work [RS'08-1] was carried out by Rui Sêca. In this work, a new Poker agent was developed named HuBot. This agent follows the probabilistic formula-based approach used in the award-winner Loki/Poki agent developed by the CPRG. It is intended to play the variant Limit Texas Hold'em, and plays best in a full ring of players.

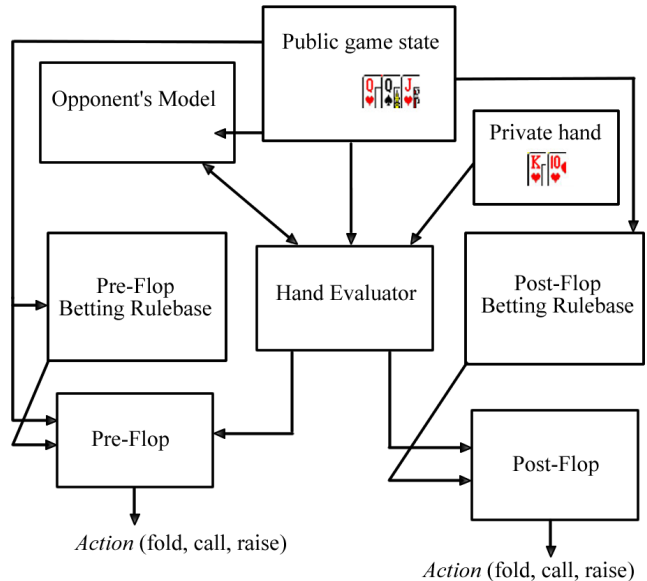

**Fig. 1 - The architectural concepts of HuBot.**

The program can be divided into three main components: pre-flop betting strategy, post-flop betting strategy and opponent modeling.

**Pre-Flop Strategy**

An initial assessment of the agent's cards is done by using Income Rate tables, which contain estimates of the expected value for each possible hand. These estimates were calculated offline in a roll-out simulation. Based on the assessment made, one strategy is selected from a fix set of rule based strategies.

**Post-Flop Strategy**

HuBot evaluates its hand comparatively to the board cards (both cards already revealed, and possible cards yet to come). This calculation also takes into account a probability distribution over the possible hands each opponent might hold. This distribution is implemented in the form of a weight table.

**Opponent Modeling**

One weight table is maintained for each opponent, and is updated after each action. This is called re-weighting, and depends on the action frequencies observed for that player (e.g. a player usually raises 20% of times in a given context, thus we infer that this player raises with the 20% best hands). The reweighting function uses linear interpolation so as to allow more flexibility to the agent's assumptions.

The action frequencies tables represent a statistical specific opponent modeling (SOM) and two tables are kept per opponent: one for the first decision in the round and another for further decisions.

**Results**

Three test scenarios were considered. In the first, the agent played against an older version of itself, five Poki agents, and two simulation-based agents, in the advanced table. HuBot managed to break even in this table, with an income rate of 0.00sb/hand, after 27,600 hands were played. The older version lost at a rate of -0.04sb/hand, as its playing style is much more predictable than the current version's.

In the second scenario, HuBot was put to play against seven un-adaptative agents (Jagbots) and one Poki, in the beginners table. HuBot's performance was the best of the table, with a steady income rate of +0.08sb/hand.

Finally, HuBot played again in the advanced table, against a version of HuBot (version 113b) without opponent modeling, and against the same other agents as before. This proved the importance of opponent modeling, as HuBot v113b showed an income rate of -0.14sb/hand, in comparison to the normal HuBot, who performed here with an income rate of +0.02sb/hand.

## 3. Learning Pre-flop Strategies in Multiplayer Tables (2008)

This work [JF'08-1] was developed by João Ferreira. It consists in determining which factors promote changes in a Poker strategy and measure their importance. Thus, this work presents a causal model of the game of Poker and so human player hands were used for game analysis. They were extracted from BWin website through the observation of live games and were used to analyze the following features of the table:

- **Position in table:** the extracted data demonstrated that players Fold more in early positions.
- **Number of players:** when the number of players is higher, the fold ratio is also higher.
- **Other player actions:** the fold ratio increases greatly when the first player raises.
- **Number of chips:** in tournaments the number of chips is a key factor and it influences the players' actions. The situation in online games differs from that of live playing.

**Results**

The results show that factors like position of the player, number of players at the table, chips and other player's actions are relevant for the strategy of the players. From these factors, the actions of the others players is the factor causing the most significant changes of strategy. From the results it is also evident that the changes in strategy are not random but indeed follow a specific pattern.

## 4. High-Level Language to build Poker Agents (2008)

This work was undertaken by Pedro Mendes [PM'08-1] and Nuno Cruz [NC'09-1] and resulted in two master theses. The main goal of the project was to create a powerful tool capable of creating Poker Agents through rules of concepts, so that any user, even without computer programming knowledge, can easily create his/her own agent.

**PokerLANG**

In this work, the first step was to create a high-level language of poker concepts: PokerLANG allows for the construction of poker agents with a "language" that normal poker players would comprehend. The language follows a format similar to the RoboCup Coach Language (Coach Unilang), a language developed to enable online coaches to change the behaviour of simulated soccer players during games in the Simulated League of the robotic soccer international competition – RoboCup.

```
<STRATEGY>::= {<ACTIVATION_CONDITION> <TACTIC>}
<ACTIVATION_CONDITION>::= {<EVALUATOR>}
<TACTIC>::= <PREDEFINED_TACTIC> | <TACTIC_NAME><TACTIC_DEFINITION>
<PREDEFINED_TACTIC>::= loose_agressive | loose_passive | tight_agressive | tight_passive
<TACTIC_NAME>::= [string]
<TACTIC_DEFINITION>::={<BEHAVIOUR> <VALUE>}
<BEHAVIOUR>::= {<RULE>}
<RULE>::= {<EVALUATOR> | <PREDICTOR>} <ACTION>
<EVALUATOR>::= <NUMBER_OF_PLAYERS> | <STACK> | <POT_ODDS> |
              <HAND_REGION> | <POSITION_AT_TABLE>
<PREDICTOR>::= <IMPLIED_ODDS> | <OPPONENT_HAND> | <OPPONENT_IN_GAME> |
              <STEAL_BET> | <IMAGE_AT_TABLE>
<ACTION>::= {<PREDEFINED_ACTION><PERC> | <DEFINED_ACTION><PERC>}
<PREDEFINED_ACTION>::= <STEAL_THE_POT> | <SEMI_BLUFF> |
                      <CHECK_RAISE_BLUFF> | <SQUEEZE_PLAY> |
                      <CHECK_CALL_TRAP> | <CHECK_RAISE_TRAP> |
                      <POST_OAK_BLUFF>
```

Fig. 2 - PokerLANG Main Definition

**Poker Builder**

An application with a simple graphical interface was created in order to support and help the users creating their Poker Lang strategies.

An agent that follows a Poker Lang strategy was also created and it showed interesting results against agents created by experts in the area.

## 5. Building a Poker Playing Agent based on Game Logs using Supervised Learning (2010)

This work [LT'10-1] was developed by Luís Filipe Teófilo and culminated in the publication of two papers [LT'11-1, LT'11-2].

The focus of this work was to verify whether is possible to analyze human game logs to produce competent Poker agents. For that reason, the HoldemML Framework was produced.

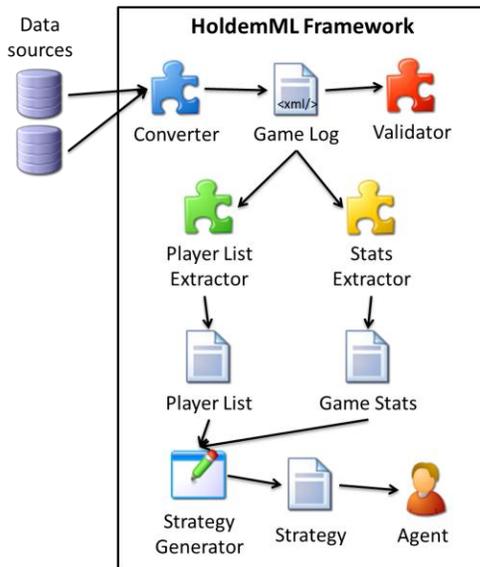

**Fig. 3. HoldemML Framework**

The HoldemML framework contains a Converter application that receives game logs from different data sources. Afterwards, it converts the game logs into a common format structure (in XML). After all the data is processed two documents are created: "Player List" – contains the list of all relevant players present in the data source – and "Game Stats" – calculates the game state (position score, effective hand strength, type of the last player, etc…) for each action. These two files are used to generate a strategy file which is used by the agent to reproduce the human strategy. The strategy file is created by applying a user-defined supervised learning algorithm.

The agent can use several strategy files at the same time and it changes the file throughout the game using a simple heuristic: when a strategy loses money for some time, it changes.

**Results**

After the implementation of the framework, three types of tests were used to validate this approach: classifier tests, behavior tests and game tests.

The classifier tests showed that the best classifier to recognize strategies in logs was a Random Forest Tree because it presents lower average error.

The behavior tests showed that generated agents have a behavior similar to the human player they are trying to imitate because they have got very similar VP%IP and aggression factor.

Finally, the game tests showed that the agents were able to outperform simple adversaries, but since they use a fixed strategy any agent with opponent modeling skills is capable of beating them. That problem was solved by mixing strategies from different human players, to confuse the opponent modeling mechanisms.

## 6. Poker Learner: Reinforcement Learning Applied to Texas Hold'em Poker (2011)

This work [NP'11-1] completed by Nuno Passos was also published as a paper [LT'12-2]. It combines pre-defined opponent models with a reinforcement learning approach. The decision-making algorithm creates a different strategy against each type of opponent by identifying the opponent's type and adjusting the rewards of the actions of the corresponding strategy. The opponent models are simple classifications used by Poker experts. Thus, each strategy is constantly adapted throughout the games, continuously improving the agent's performance. In light of this, two agents with the same structure but different rewarding conditions were developed and tested against each other and other agents.

**Approach**

The agents were designed with a Q-Table containing the state-action pairs. The state ($\sigma$) is defined as:

- **G**: A value representing a pair of cards that compose the player's hand. This is useful since many hands have the same relative value (e.g. {2♣, 4♥} and {2♦, 4♣}).
- **P**: The player's seat on the table (big-blind or small-blind).
- **T**: A value representing the opponent type (Tight Aggressive, Tight Passive, Loose Aggressive and Loose Passive).
- **A**: A value representing the last action before the agent's turn (Call, Raise).

Each state has a direct correspondence to tuple (**C** – call weight, **R** – raise weight) as described by the following equations.

$$\sigma(G,P,T,A) \to (C,R) : C + R \leq 1;$$
$$G \in \{0 \ldots 127\}; P \in \{'Big,' Small'\}; T \in$$
$$\{'TA','TP','LA','LP'\}; \quad (2)$$
$$A \in \{'Call','Raise'\}; C,R \in [0,1]$$

The Q-Table is initially empty and the weights are filled up with random numbers as there is need for them. The value of the weights stabilizes as the games proceed, so as to choose the option which maximizes profit. However convergence to stable weight values is not guaranteed because the game state to action mapping may not be sufficient to fully describe the defined opponent types.

When the agent plays, it searches the Q-Table to obtain the values of C and R so as to decide on the action to take. After retrieving these values, a random number ($N \in [0,1]$) is generated. The probability of choosing an action is:

$$Action = \begin{cases} Call, N \in [0, C] \\ Raise, N \in ]C, C+R] \\ Fold\ otherwise \end{cases}$$

The flowchart describes the complete process of update and us-age of the Q-Table.

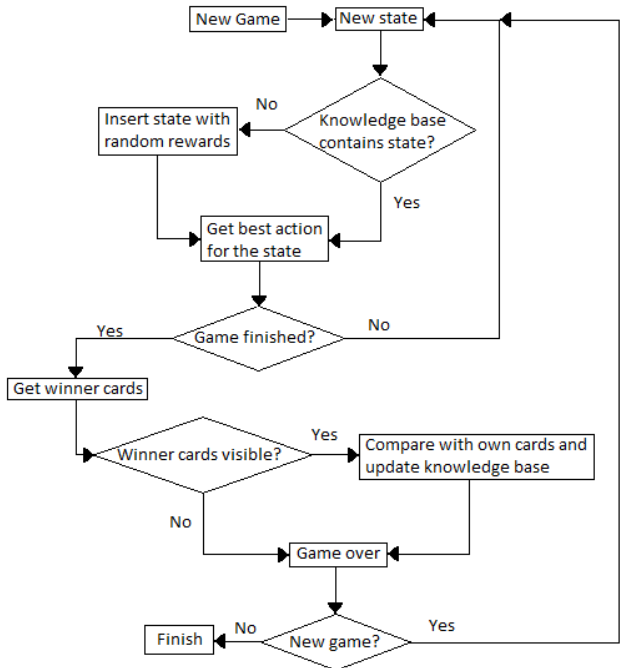

**Fig. 4 - Structure of the agent's behavior**

Two agents with this structure were implemented: WHSLearner and WHLearner. The only difference between them resides on the reward calculation. Whilst WHSLearner updates the rewards based on the evaluation of the adequacy of the decision, WHLearner considers the actual outcome of the game. The next table shows how C and R variables are updated.

**Table. 1 – Decision matrix for WHSLearner WHLearner agents**

| Agent | | Agent Action | | |
|---|---|---|---|---|
| WHS Learner | WH Learner | Fold | Call | Raise |
| Good Choice | Game Won | C↓, R↓ | C↑, R↑↑ | C↓, R↑↑ |
| Bad Choice | Game Lost | C↑, R↑↑ | C↓, R↓↓ | C↑, R↓↓↓ |

**Results**

Results showed that this approach is a valid starting point to create a complete Texas Hold'em agent, since the agent outperformed every opponent in all experiments. Another important conclusion can be extracted from the differences between the performance of WHSLearner and WHLearner. In most experiences, WHSLearner performed better, which means that rewarding good decisions may be a better approach than rewarding good outcomes in reinforcement learning algorithms.

## IV - Current Research

This section briefly describes current research works at LIACC about Computer Poker. This is mostly a summary of the Ph.D. work presently being developed by Luís Filipe Teófilo.

### General Approach

The Ph.D. research project is currently named "Development of competitive Texas Hold'em Agents with adaptive strategies to high-level opponent models". It consists on the development of software modules that will interact as depicted in the figure below. Each module corresponds to the completion of one of the Ph.D. thesis goals.

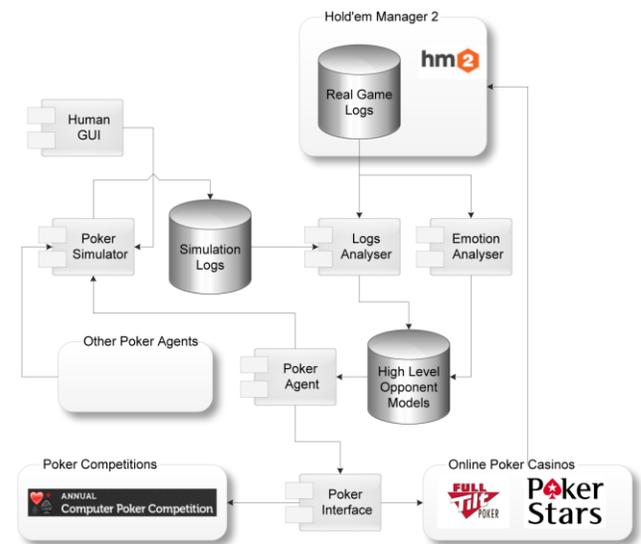

**Fig. 5 - Research work global architecture**

In the figure it is possible to identify the modules to be implemented (represented as UML components) as well as external modules that interact with those. Below follows a brief description of each module that constitutes the global architecture of the Ph.D. research work plan:

- Poker Simulator – a new simulation system to support Computer Poker research.
- Simulation Logs – the simulation logs produced by the new Poker Simulator.
- Human GUI – a GUI that will communicate with the simulator in order to allow human players to play against Poker agents.
- Logs Analyzer – this tool is responsible for creating Poker player profiles (opponent models) from game logs.
- Emotion Analyzer – emotion modeling capabilities for Poker agents will be created to enable agents to obtain advantage in the game by exploring weaknesses related with the emotional state of the human opponents.
- High Level Opponent Models – this is a database of opponent models which associates complex strategies to combinations of opponent characteristics.
- Poker Agent – several agents will be produced based on improvements on the current state of the art as well as new methodologies.
- Poker Interface – a bridge between Poker agents and human players (Poker Bot). This application will allow agents to easily play against human players in real money games.
- Hold'em Manager 2 – this is an external application which records and manages all game logs of installed Poker clients. It also displays real time opponent evaluation.
- Poker Competitions – these competitions take place between Poker agents and are useful to assess advances on the current state of the art.
- Online Poker Casinos – this is software which allows Poker players to play online.

## A Simulation System to Support Computer Poker Research

The competitiveness of Poker agents is typically measured through simulation systems. However, current systems do not provide an adequate toolset for assessing the agents' capabilities since they were built to play and not specifically for research. For that reason, a new simulation system was created [LT'12-1]. This system considers the bankroll management component of the game, allowing the evaluation of the agents' survivability between games, with limited initial recourses (tournaments). The system also supports assessing agents in several game modes like an evolutionary environment, ring games and cash games. The figure bellow presents the global architecture of the new simulator.

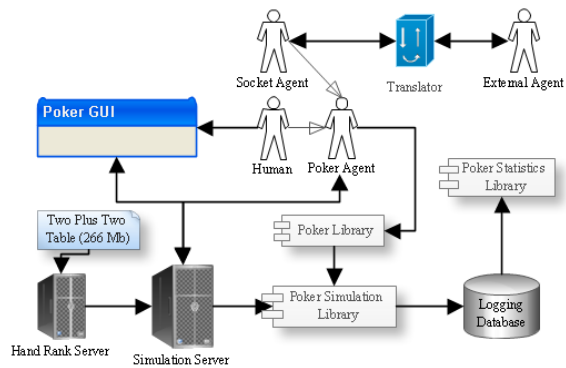

**Fig. 6 - LIACC Poker Simulator Architecture**

The simulator will support further research into Computer Poker, thus fomenting the creation of an autonomous agent that considers all game components.

### High Level Actions in Poker

Most Poker agents simply choose a single action (Call, Raise or Fold) after processing the current game state and the game moves history. In this work there is an attempt to map the processing into round-oriented high level actions (like human players do) or sequences of actions. The full set of possible actions is yet to be decided, but some examples could be: "Check Raise", "Raise Call" or "Semi Bluff".

### Emotions in Poker (Tilt analysis)

Tilt is an emotional state in a game of Poker, based on emotional confusion or frustration that affects the player's behavior in the game, which causes the player to use a less optimal strategy than usual. Tilt is usually experienced after big losses of money in Poker, but large gains can also affect the strategy of a human player since they might promote overconfidence, which can result in careless play.

This work consists in developing mechanisms for Poker agents to detect possible tilts in human opponents. By detecting tilts, the agent will likely improve the results against human players because it takes advantage of their emotional state. Initially the methodology will be tested against agents that simulate emotions and then tests will be conducted with human players. The aim is to determine to what extent an agent that detects emotions can improve its performance in Poker. Tests with human players will provide a more accurate form of validation of this approach as well as the validation of the agents that simulate emotions in Poker.

## V - Conclusions

This paper summarized the main methodologies followed by LIACC's researchers. Despite the number of research works about Poker it is important to note that LIACC could benefit from an increase in communication with other Poker research groups to further improve the quality of Computer Poker research. The effects of the present lack of communication were felt on publications which were unaware of recent methodologies such as Counterfactual Regret Minimization or the Monte Carlo Search Tree algorithm.

**Acknowledgments.** Luís Filipe Teófilo would like to thank Fundação para a Ciência e a Tecnologia for supporting his work by providing a Ph.D. Scholarship SFRH/BD/71598/2010.